%% file: main.tex
\documentclass[runningheads]{llncs}

\usepackage{eccv}

\usepackage{eccvabbrv}

\usepackage{graphicx}
\usepackage{booktabs}

\usepackage[accsupp]{axessibility}  

\usepackage[breaklinks,colorlinks,citecolor=eccvblue]{hyperref}

\usepackage{orcidlink}

\usepackage{multirow}
\usepackage{multicol}
\usepackage[table]{xcolor}
\usepackage{pifont}
\usepackage{cuted}
\usepackage{enumitem}
\definecolor{cvprblue}{rgb}{0.21,0.49,0.74}

\newcommand{\ourtask}{\textsc{CoCo-IR}\xspace}
\newcommand{\ourmodel}{\textsc{TIE}\xspace}

\begin{document}

\title{\ourtask: Contextual Composed Image Retrieval}

\titlerunning{\ourtask: Contextual Composed Image Retrieval}

\author{Shengcao Cao\inst{1,2} \and Tanmaya Shekhar Dabral\inst{2} \and Zhongli Ding\inst{2} \and \\ Madhuri Shanbhogue\inst{2} \and Kaifeng Chen\inst{3} \and Zhe Li\inst{2} \and Mojtaba Seyedhosseini\inst{2} \and \\ Yu-Xiong Wang\inst{1} \and Liang-Yan Gui\inst{1}}

\authorrunning{S.~Cao et al.}

\institute{University of Illinois Urbana-Champaign \and Google DeepMind \and OpenAI. Work done at Google DeepMind.}

\maketitle

\input{sec/0_abstract}
\input{sec/1_intro}
\input{sec/2_related}
\input{sec/3_method}
\input{sec/4_experiments}
\input{sec/5_conclusion}

%
%
\bibliographystyle{splncs04}
\bibliography{main}

\input{sec/X_suppl}

\end{document}

%% file: sec/0_abstract.tex
\begin{figure}
    \centering
    \includegraphics[width=\columnwidth]{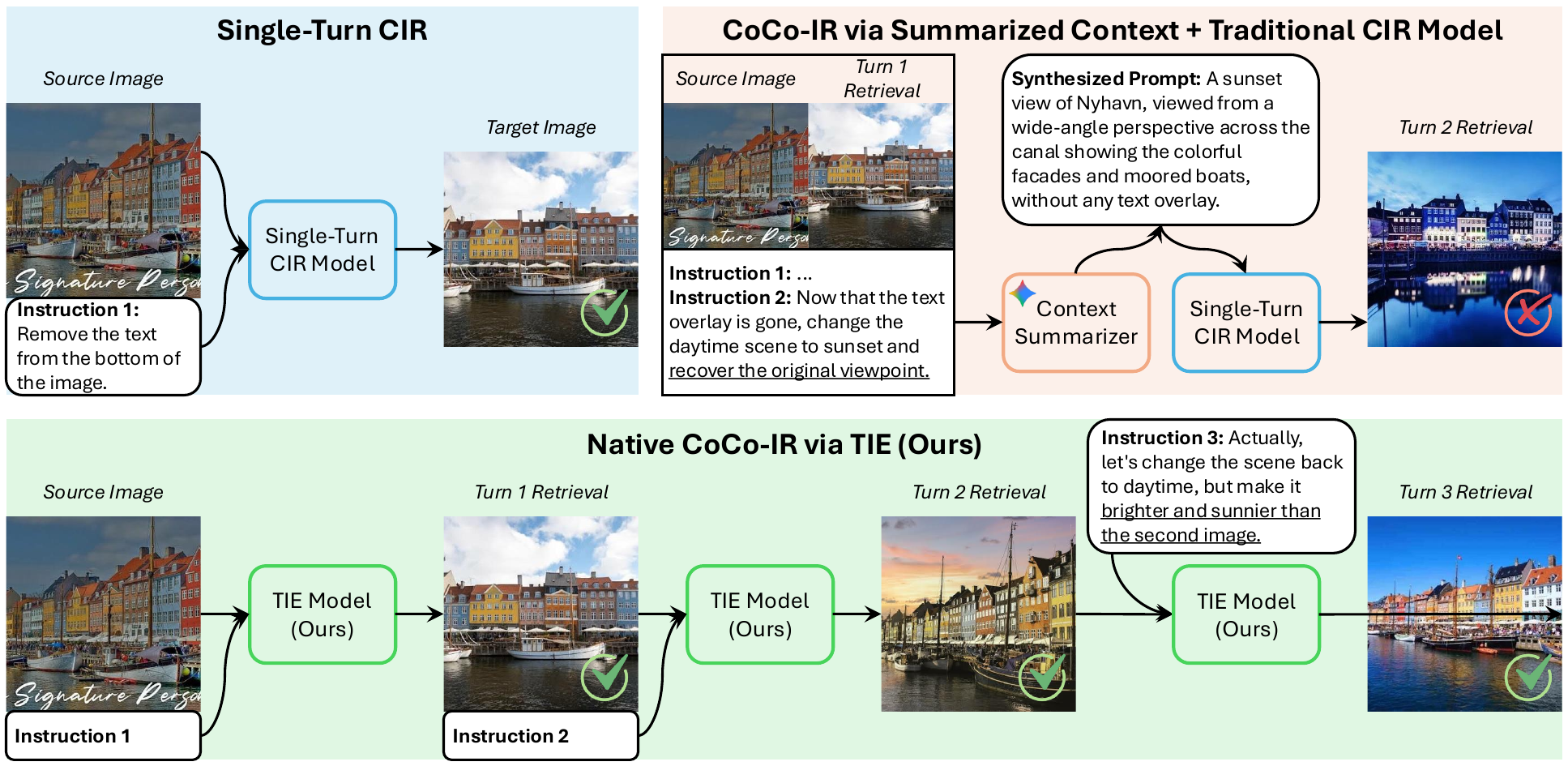}
    \caption{{Comparison between conventional single-turn Composed Image Retrieval (CIR) and our novel Contextual Composed Image Retrieval (\ourtask) task, achieved via our proposed \ourmodel framework.} \textbf{(Top Left)} A standard single-turn CIR model successfully processes a simple, isolated instruction. \textbf{(Top Right)} Traditional CIR models are architecturally unable to process multi-turn context with interleaving images and text. Even with a strong external model (\eg, Gemini-2.5-Pro) that summarizes the entire history into a single prompt, context information may inevitably be compressed, leading to a failed image retrieval. \textbf{(Bottom)} Our proposed \textbf{\ourmodel} natively processes an evolving context across multiple turns, enabling seamless comprehension of context-dependent instructions that refer to specific past images (\eg, ``original viewpoint'' and ``second image''), successfully completing the entire contextual retrieval trajectory within one unified framework.}
    \label{fig:teaser}
\end{figure}

\begin{abstract}
Current instruction-based image retrieval systems are powerful but limited to single-turn interactions, failing to capture the iterative nature of complex, real-world visual searches. To overcome this limitation, we introduce \textbf{Contextual Composed Image Retrieval (\ourtask)}, a novel task that enables users to progressively refine search results through interactions. We address this new task by proposing a new model based on a Large Multimodal Model (LMM) that functions as a context-aware reasoner for \ourtask. Our model interprets the entire interaction history to generate \textbf{Transformable Image Embeddings (\ourmodel)} that evolve across turns. To fuel the model training without expensive human annotations, we develop a fully autonomous, scalable data engine that leverages LMMs to generate high-quality contextual retrieval data, and uses model-guided verification to mine challenging hard negatives. Extensive experiments demonstrate that our approach establishes new state-of-the-art performance: We achieve 39.4 mAP@5 on the challenging single-turn benchmark CIRCO; furthermore, on our new \ourtask benchmark, our model maintains robust performance with 44.1 R@1 on 4-turn dialogues, dramatically outperforming existing methods (28.2 4-turn R@1) that fail to handle multi-turn context. Project page: \url{https://CoCo-IR.github.io}.
\end{abstract}

%% file: sec/1_intro.tex
\section{Introduction}
\label{sec:intro}

The quest for effective visual search has rapidly evolved from simple keyword queries to sophisticated instruction-based systems. A prominent example is Composed Image Retrieval (CIR)~\cite{vo2019composing}, where a user provides a source image $I_\text{src}$ and a natural language instruction $T$ to retrieve a target image $I_\text{tgt}$ that reflects the desired transformation (\eg, ``change the car to red''). This paradigm, powered by large-scale web data~\cite{schuhmann2022laion, zhai2022scaling, chen2023pali} and multimodal encoders~\cite{radford2021clip, zhai2023sigmoid, yu2022coca}, has seen significant progress through improved modality fusion~\cite{saito2023pic2word, gu2024language} and foundation model reasoning~\cite{karthik2024vision}.

However, a fundamental limitation of classical CIR is its restriction to \emph{isolated, single-turn} queries. By processing only one query at a time, these models fall short of the natural, iterative, and exploratory nature of real-world visual search. In practice, a user's intent is often too complex or evolving to be captured in a single command. For instance, a user might want to \emph{progressively refine} their search based on intermediate results, \emph{explore a sequence of related but distinct images}, where the intermediate discoveries are just as important as the final target, or even \emph{change their mind} partway through the process (as illustrated in Fig.~\ref{fig:teaser}). Existing CIR models cannot support this interactive dynamic; any refinement, exploration, or pivot forces the user to start over. Instead of building upon established visual context, users must attempt to craft a new, complex prompt from scratch, highlighting a critical gap between current capabilities and real-world user needs.

To bridge this gap, we introduce a new task: \textbf{Contextual Composed Image Retrieval (\ourtask)}. We formulate instruction-based retrieval as an interactive dialogue, where users progressively refine their search over multiple turns. In this framework, the model must interpret each new instruction $T_m$ relative to the \emph{entire} interaction history $H_m = (I_0, T_1, I_1, \dots, T_{m-1}, I_{m-1})$ that includes the initial source image $I_0$, previous instructions $T_1,\dots,T_{m-1}$, and intermediate images $I_1,\dots,I_{m-1}$. This contextual setting is inherently more flexible and powerful, enabling users to decompose complex visual search goals into a sequence of simpler, more manageable steps, making the search process more intuitive and effective.

This new \ourtask task demands a paradigm shift in modeling. As our experiments (Tab.~\ref{tab:multi_turn}) demonstrate, prior methods fail to support \ourtask because they cannot natively encode a sequence of interleaved images and instructions, \emph{even with a powerful external model that can summarize the context}. To solve this, we propose \textbf{Transformable Image Embedding (\ourmodel)}, which unifies multimodal comprehension and embedding generation into a single Large Multimodal Model (LMM)~\cite{team2025gemma}. By enabling end-to-end optimization, \ourmodel inherently understands the evolving context, yielding substantially stronger performance.

To effectively process a sequence of multimodal turns, we enhance the LMM with two key architectural designs. First, in contrast to prior embedding models that extract the hidden state of the final text token, which is inherently biased toward its immediate semantic predecessors, we introduce a specialized \texttt{\textlangle EMB\textrangle} token. \texttt{\textlangle EMB\textrangle} acts as a dedicated global information bottleneck, forcing the model to aggregate the full interaction history into a compact, query-aware embedding. Second, to handle the structure of multi-turn interactions, we design a hybrid attention mechanism by employing bidirectional attention \emph{within} each turn for deep multimodal fusion, and causal attention \emph{across} turns to respect the temporal flow of the interaction. The resulting transformable image embeddings accurately capture the cumulative semantic transformation specified by the entire dialogue, not just the most recent instruction.

Fueling this \emph{context-aware} model requires a large-scale multimodal long-context dataset, which, to our knowledge, does not exist yet. Manually annotating such data is prohibitively expensive. We address this by developing a \emph{scalable, LMM-powered data engine} with two key innovations. First, for instruction generation, we anchor the process with clustered real images and leverage \textbf{self-reflection}, where an LMM generates a transformation instruction and then scores its own quality and ambiguity, allowing us to filter for high-quality data. Second, to learn fine-grained distinctions, we employ LMMs as \textbf{verifiers} for mining challenging hard negatives, which are candidate images that are visually similar to the target but fail the instruction (Fig.~\ref{fig:negative}). This autonomous data engine proves to produce data of high quality in our human verification. Additionally, as demonstrated in single-turn results, we achieve state-of-the-art performance using $14\times$ fewer training samples than prior work~\cite{zhang2024magiclens}.

In summary, our key contributions are:
\begin{itemize}[leftmargin=*, noitemsep, nolistsep]
    \item We introduce \ourtask, a new task for \emph{multi-turn contextual composed image retrieval} that better reflects real-world image search.
    \item We propose \ourmodel, an LMM-based architecture that generates \emph{context-aware, transformable image embeddings} that capture the full dialogue history.
    \item We develop a scalable data engine that leverages LMM-driven self-reflection and hard-negative verification to create the first large-scale \ourtask dataset.
    \item We demonstrate state-of-the-art performance on both single-turn CIR benchmarks (\eg, 39.4 mAP@5 on CIRCO~\cite{baldrati2023zero}) and our new multi-turn benchmark (\eg, a robust 44.1 R@1 on 4-turn context where prior methods collapse).
\end{itemize}

%% file: sec/2_related.tex
\section{Related Work}
\label{sec:related-work}

\noindent\textbf{From Multimodal Encoders to LMMs.}
Early successes in vision-language pre-training come from dual-encoder models that align image and text representations in a joint space~\cite{radford2021clip, yu2022coca, li2021align, zhai2022scaling, schuhmann2022laion}. These models excel at zero-shot cross-modal retrieval but lack the deep fusion necessary to understand complex instructions. Recent efforts propose large multimodal models (LMMs)~\cite{li2022blip, li2023blip2, alayrac2022flamingo, liu2023visual, liu2024improved, li2024llavanext-strong} that integrate vision encoders with large language models, enabling more sophisticated reasoning and instruction-following capabilities~\cite{chen2023can}. Our work builds on this LMM paradigm, as the ability to interpret and understand multiple interactive rounds is fundamental to our contextual retrieval task.

\noindent\textbf{Composed Image Retrieval (CIR).}
Our work directly extends composed image retrieval~\cite{vo2019composing} (CIR), where a query consists of a source image and a modifying text instruction. A significant body of work has focused on this single-turn task, designing lightweight modality transformers~\cite{saito2023pic2word, gu2024language, baldrati2023zero} or leveraging strong reasoning from foundation models~\cite{karthik2024vision}. To address inherent data bottlenecks, prior work has established static benchmarks~\cite{wu2021fashion, liu2021image, baldrati2023zero} and increasingly relied on synthetic data generation pipelines~\cite{brooks2023instruct, gu2024compodiff, zhang2024magiclens, zhou2025megapairs}. Our proposed data engine differs significantly from these prior efforts. Specifically, we advance beyond single-step retrieval by generating \emph{multi-turn} interactions, integrating \emph{hard negative mining} to improve embeddings, and applying rigorous \emph{filtering and quality control} to ensure context coherence.
IRR~\cite{wei2023conversational} and MAI~\cite{chen2025mai} extends CIR to a scenario where the user provides a reference image only at the first turn, and subsequent turns are text-only refinements. In contrast, our formulation supports \emph{interleaved image-text input at every turn}, allowing users to introduce new visual context mid-session. Both IRR and MAI are also limited to the fashion domain, while ours is open-domain using a general-purpose LMM.

%% file: sec/3_method.tex
\section{Approach}
\label{sec:method}

In this section, we first formalize our new task, Contextual Composed Image Retrieval (\ourtask). We then describe the details of our model architecture, training procedure, and finally our scalable data engine.

\subsection{Task Formulation: From Single-Turn to Multi-Turn Contextual Retrieval}
\label{sec:task}

We begin by formally introducing the task of \textbf{Composed Image Retrieval (CIR)}~\cite{vo2019composing}, which serves as the foundation for our multi-turn task. In the standard CIR setting, the goal is to retrieve a specific target image $I_\text{tgt}$ from a large corpus of images $\mathcal{C}$. The query consists of a source image $I_\text{src}$ and a free-form natural language instruction $T$ that describes a desired transformation or a semantic relationship between the source and target images. A model is trained to learn a function that maps the composed query $(I_\text{src}, T)$ and the target image $I_\text{tgt}$ into a shared embedding space, such that the distance between the query embedding and the correct target embedding is minimized.

While powerful, this single-turn paradigm does not capture the iterative nature of complex real-world visual searches. We therefore extend this paradigm to a \textbf{multi-turn, interactive setting}. This framework allows a user to progressively refine their search through an interactive dialogue, reusing the context from previous interactions to inform subsequent queries.

A multi-turn retrieval session begins similarly to the single-turn case. At the first turn ($t=1$), the user provides an initial source image $I_0$ and a text instruction $T_1$. The retrieval system's task is to return the corresponding target image $I_1$, among up to top-$k$ predictions, from the corpus $\mathcal{C}$. The key difference emerges in subsequent turns. For any turn $t > 1$, the user provides a new instruction $T_t$ that can \emph{refer to any element in the preceding interaction history}, $H_t = (I_0, T_1, I_1, \dots, T_{t-1}, I_{t-1})$. The model must then interpret $T_t$ in the context of this history to retrieve the next target image, $I_t$. This process can continue for an arbitrary number of turns, enabling the decomposition of a complex search intent into a sequence of simpler, context-aware steps.

\noindent\textbf{Evaluation Metric.}
To evaluate performance in this new interactive setting, we must bridge the gap between dynamic user behavior and the requirements of standardized evaluation. As shown in Fig.~\ref{fig:teaser}, intermediate images in a interactive search often hold independent value or serve as critical stepping stones. In a real-world interactive search, if a user does not immediately see their desired intermediate image in the top-$k$ results, they naturally adapt their subsequent instructions based on what was actually retrieved. However, to construct a rigorous and reproducible benchmark, evaluation trajectories (including all intermediate target images and their corresponding sequential instructions) must be fixed and predefined. Consequently, if the model fails to retrieve the intended ground-truth target within the top-$k$ results at any turn, the retrieval trajectory diverges from the benchmark's established path. When this happens, predefined subsequent instructions that rely on that specific missing context (\eg, in Fig.~\ref{fig:teaser}, commanding the system to ``make it brighter and sunnier than the \emph{second image}'') immediately become ambiguous or incoherent.

Because we cannot simulate on-the-fly user adaptation, our evaluation must measure the model's ability to independently sustain this exact, unbroken chain of correct retrievals. Furthermore, because real-world multi-turn retrieval is often an exploratory process, intermediate discoveries are just as important as the final destination. Therefore, rather than merely evaluating the accuracy of the final result, we introduce the \textbf{$m$-turn Recall@$k$} metric to explicitly validate the entire trajectory. We define an $m$-turn query as a success if and only if the ground-truth target image for \emph{every} turn is successfully retrieved within the top-$k$ results for that respective turn. Formally, let the sequence of ground-truth target images for an $m$-turn interaction be $(I_1^*, I_2^*, \dots, I_m^*)$, and let $\mathcal{R}_t^k$ be the set of top-$k$ images retrieved by the model at turn $t$, given the ground-truth history up to step $t$: $(I_0, T_1, I_1^*, \dots, I_{t-1}^*, T_t)$. The entire retrieval is marked as correct if and only if $\bigwedge_{t=1}^{m} (I_t^* \in \mathcal{R}_t^k)$. While this cumulative success criterion is strict, it is a necessary and logical design choice for a static benchmark: it penalizes off-trajectory compounding errors and ensures the model is rewarded only when it successfully supports the user's entire exploratory journey. Ultimately, $m$-turn Recall@$k$ provides a robust measure of a model's true contextual understanding over successive interactions.

\begin{figure*}[t]
    \centering
    \includegraphics[width=\columnwidth]{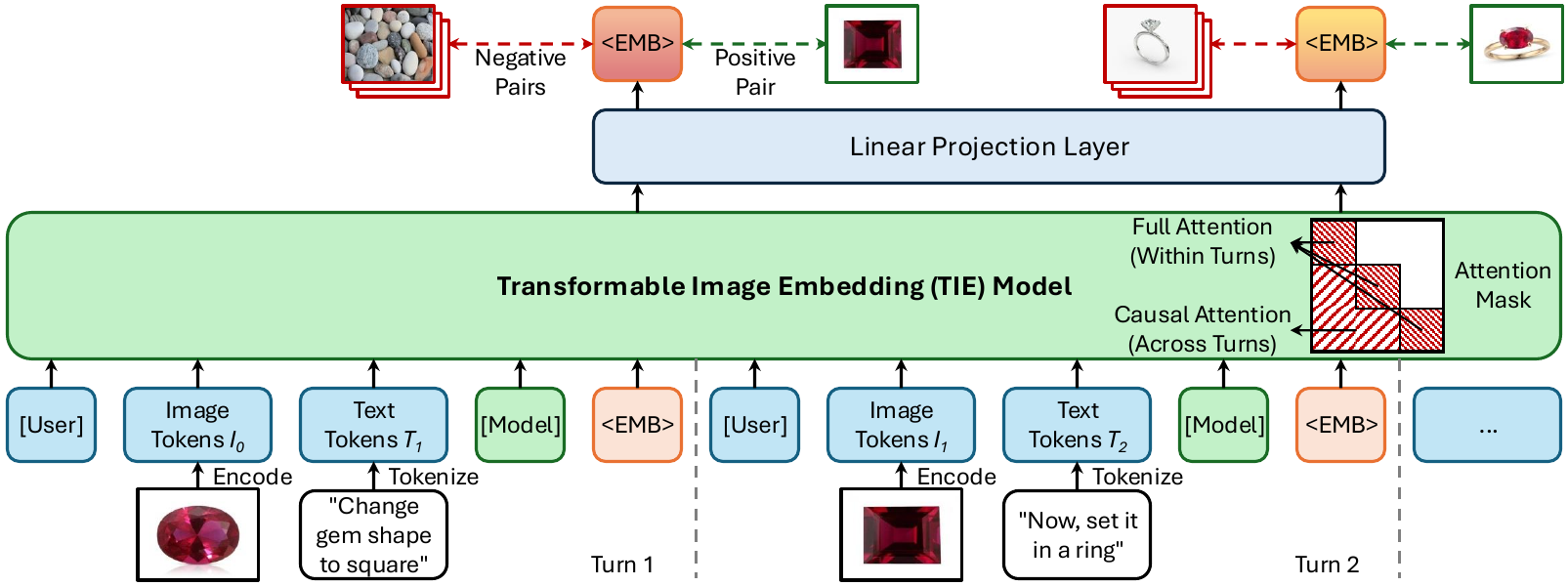}
    \caption{\textbf{Overview of our Transformable Image Embedding (\ourmodel) model.} \ourmodel processes the multi-turn context as a unified token sequence (bottom), formatting the visual and textual inputs into an interleaved dialogue. A key architectural component is the special \texttt{\textlangle EMB\textrangle} token at the end of each turn, which acts as a global bottleneck to summarize the full preceding context. As illustrated by the hybrid \textbf{Attention Mask} (middle right), the model employs full (bidirectional) attention within each turn for deep multimodal fusion, and causal attention across turns to respect the temporal flow. Finally, the embedding generated by the \texttt{\textlangle EMB\textrangle} token passes through a linear projection layer and is trained with a contrastive learning objective (top). This loss pulls the embedding toward the positive target image (green dashed arrows) while pushing it away from negative samples in the batch (red dashed arrows).}
    \label{fig:arch}
\end{figure*}

\subsection{Model Architecture and Training}
\label{sec:model}

Our Transformable Image Embedding (\ourmodel) model is built upon a Large Multimodal Model (LMM)~\cite{team2025gemma}, which is not originally designed for an embedding task. We therefore enhance its architecture and training objective specifically for the task of \ourtask, as illustrated in Fig.~\ref{fig:arch}.

\noindent\textbf{Model Architecture.} The model is designed to process a sequence of multimodal, interactive turns. In each turn, user inputs, which can be text, images, or both, are encoded into a unified token sequence. Text is processed by a standard tokenizer, while images are encoded into token embeddings using an image encoder~\cite{zhai2023sigmoid}. For a query turn, both the image and text instruction are provided to the model. For encoding a target image, only the image is provided as input.

A key modification to the base LMM is our method for generating embeddings. Instead of training the model to autoregressively predict the next token, we introduce a special \texttt{\textlangle EMB\textrangle} token. When it is the model's turn to produce an embedding, we feed this special token as input, prompting the model to summarize the entire interactive history seen so far into a single, compact representation. The output embedding is derived by passing the last-layer hidden state corresponding to this \texttt{\textlangle EMB\textrangle} token through a final linear projection layer. Compared to the previous ``last token'' solution~\cite{jiang2024e5} that is biased toward immediate predecessors, this specialized \texttt{\textlangle EMB\textrangle} token acts as a dedicated global information bottleneck, encouraging better history aggregation (Tab.~\ref{tab:ab-model}).

To facilitate better integration of information within a query, we also modify the transformer's attention mechanism. While the attention across different turns remains causal to preserve the temporal sequence, we employ bidirectional (full) attention \textit{within} each turn. This allows the image tokens and text tokens of a query to freely attend to each other, resulting in a more holistic representation.

\noindent\textbf{Contrastive Training.} We train our model using a contrastive learning objective based on the InfoNCE loss~\cite{oord2018representation, radford2021clip}. The goal is to learn an embedding space where a query is close to its corresponding target image while being distant from all other negative images.

The core of our objective is the loss for a single query at a specific turn, $\mathcal{L}_{t,i}$. For the $i$-th sample at turn $t$, the model generates an embedding $q_{t,i}$ from the interactive history $H_{t,i}$ and the new instruction $T_{t,i}$. This query embedding is contrasted against the embedding of the positive target image for that turn, $p_{t,i}$ (extracted from image $I_{t,i}$).

We define a comprehensive set of candidate embeddings $\mathcal{C}_e$ for the entire batch. This set includes all target image embeddings $\{p_{k,j}\}$, all initial source image embeddings $\{s_j\}$, and all mined hard negative embeddings $\{h_{k,j}\}$ across all $n$ samples and their $m$ turns. The positive target $p_{t,i}$ is one of the embeddings within this set $\mathcal{C}_e$.

The loss for the query $q_{t,i}$ is defined as:
\begin{equation*}
\mathcal{L}_{t,i} = -\log \frac{\exp(\text{sim}(q_{t,i}, p_{t,i})/\tau)}{\sum_{e \in \mathcal{C}_e} \exp(\text{sim}(q_{t,i}, e)/\tau)}.
\end{equation*}

In this formulation, $\text{sim}(\cdot, \cdot)$ denotes cosine similarity and $\tau$ is a fixed temperature hyperparameter. The numerator aligns the query with its correct target. The denominator $\sum_{e \in \mathcal{C}_e} \exp(\text{sim}(q_{t,i}, e)/\tau)$ serves as the normalization term, summing over the positive pair and all potential negative pairs in the batch, effectively pushing the query away from all incorrect candidates. Finally, the overall loss is computed by averaging over all samples and turns.

\subsection{Scalable LMM-Powered Data Engine}
\label{sec:data}

A key challenge in training contextual retrieval models is the lack of large-scale annotated data. To overcome this, we design a fully autonomous, LMM-powered data engine that generates high-quality single-turn and multi-turn training data. This engine operates in a series of stages, from discovering image pairs to mining hard negatives and constructing interaction chains, creating a scalable, model-guided flywheel. We include more specific details of the data engine in the supplementary material.

\noindent\textbf{1. Discovering Image Pairs with Transformations.}
We begin by discovering a large set of image pairs from web images that exhibit meaningful, describable transformations. We employ two complementary strategies in parallel:
\begin{itemize}
    \item \textbf{Feature-Based Discovery:} We extract visual features of images using a pretrained image encoder~\cite{radford2021clip, he2016deep}, and group them into clusters. To encourage learning transformations across different visual concepts, each image is associated with its two nearest clusters. Within these expanded clusters, we identify image pairs whose feature similarities fall within a specific range, filtering out pairs that are either too dissimilar to be related by a simple instruction or nearly identical. To ensure diversity, we set a maximum number of pairs to sample from each cluster.
    \item \textbf{Metadata-Based Discovery:} Inspired by MagicLens~\cite{zhang2024magiclens}, we also leverage the co-occurrence of images on the same web page as a strong signal of semantic relatedness. We group images by their source URLs and propose pairs that appear together. These pairs are then filtered by a wider feature similarity threshold to capture a broader range of semantic relationships.
\end{itemize}

\begin{figure}[t]
    \centering
    \includegraphics[width=0.8\columnwidth]{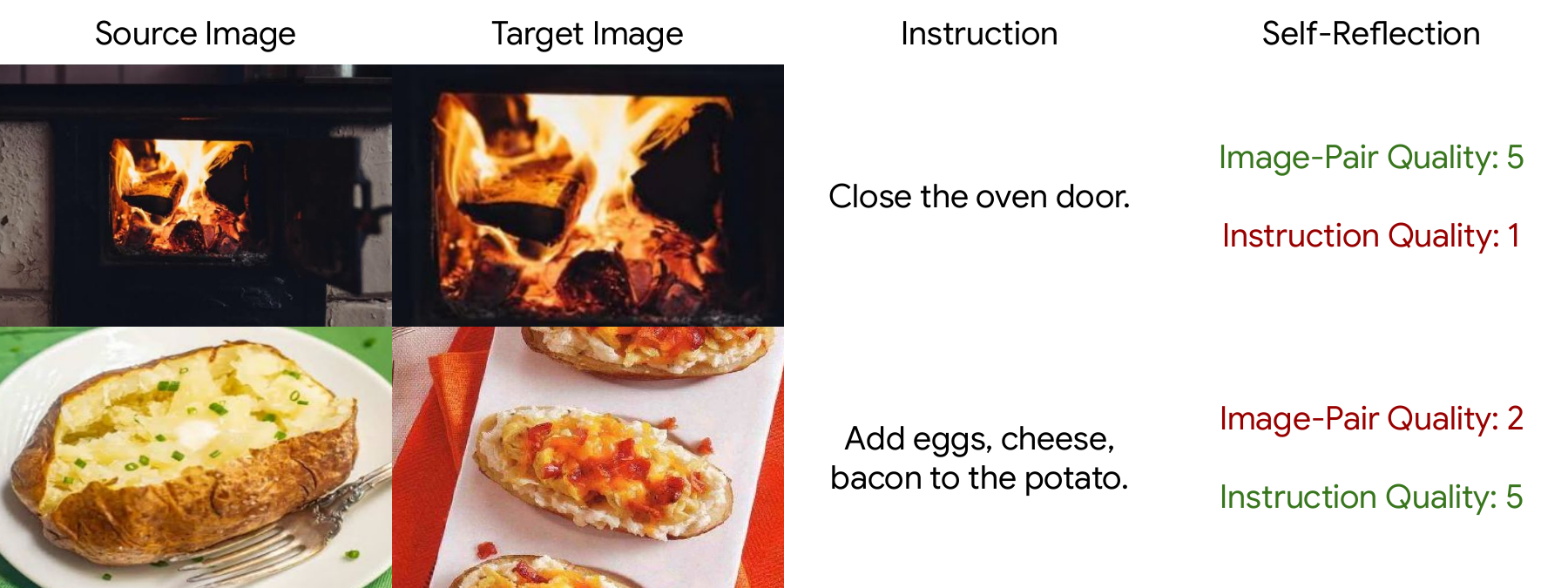}
    \caption{
        \textbf{Self-reflection during instruction synthesis.} As part of our data engine, an LMM generates a transformation instruction for a given image pair and then performs self-reflection. It assigns two scores for image-pair quality and instruction quality, respectively. The top example shows a good pair (Quality: 5) but a poor, unrelated instruction (Quality: 1). The bottom example shows a clear instruction (Quality: 5) but a poor image pair (Quality: 2), as too many visual elements are altered. This two-dimensional scoring allows us to filter for high-quality data by selecting samples where both scores are high.}
    \label{fig:self_reflect}
\end{figure}

\noindent\textbf{2. Synthesizing Instructions with Self-Reflection.}
For each discovered image pair, we prompt an LMM (\eg, Gemini 2.5~\cite{team2025gemini}) to generate a corresponding text instruction. The LMM is tasked to first perform a detailed visual comparison of the source and target images. Based on this analysis, it formulates a concise instruction that describes the transformation. Crucially, the process includes a self-reflection step: the LMM scores the quality of both the image pair itself (\textit{``Is the transformation clear and simple?''}) and its own generated instruction (\textit{``Is the instruction unambiguous and effective?''}). This two-dimensional scoring allows us to filter for high-quality data (see examples in Fig.~\ref{fig:self_reflect}).

\noindent\textbf{3. Post-Processing and Data Splitting.}
We process the generated data to create distinct training and evaluation sets. To rigorously test our model's generalization capabilities, we split the data at the image cluster level, ensuring that all images from a held-out subset of clusters are unseen during training. Furthermore, we apply different quality filters based on the LMM's self-reflection scores. For the training set, we use a more relaxed threshold to maintain data diversity, while for the evaluation set, we enforce the highest possible scores to ensure benchmark quality.

\begin{figure*}[t]
    \centering
    \begin{minipage}[b]{0.49\textwidth}
        \centering
        \includegraphics[width=\textwidth]{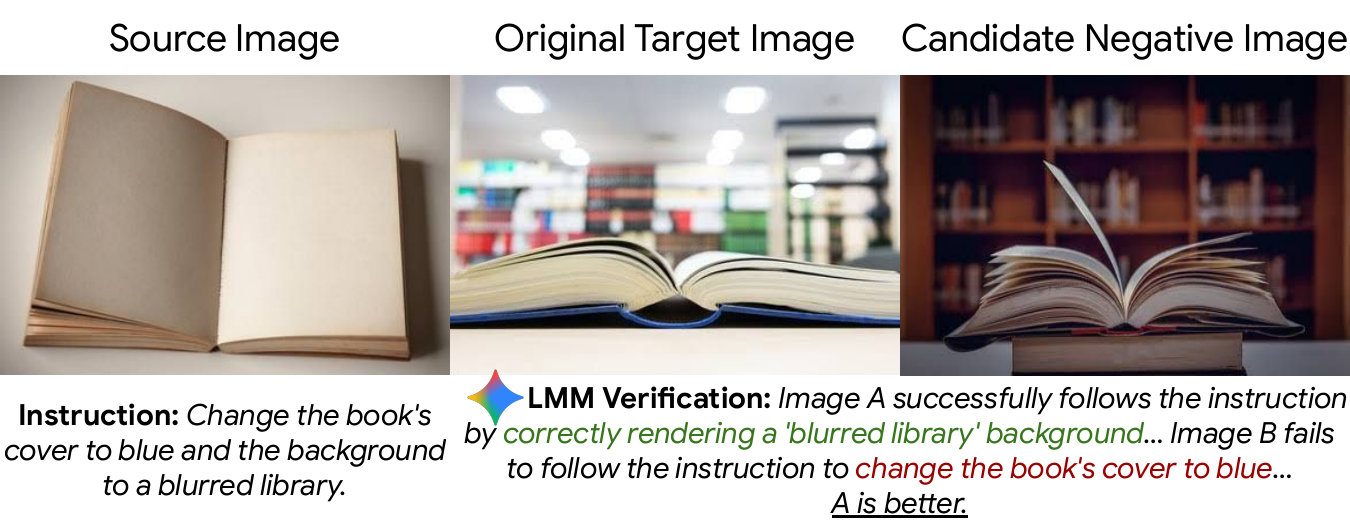}
        \caption{\textbf{Verification of hard negative samples.} Hard negatives are images that are visually and semantically similar to the ground-truth target (middle) but fail to perfectly satisfy the instruction (left). Here, the candidate negative (right) correctly matches the ``blurred library'' background but fails the ``change the book's cover to blue'' command. Our data engine uses an LMM to automatically verify this distinction.}
        \label{fig:negative}
    \end{minipage}\hfill
    \begin{minipage}[b]{0.49\textwidth}
        \centering
        \includegraphics[width=\textwidth]{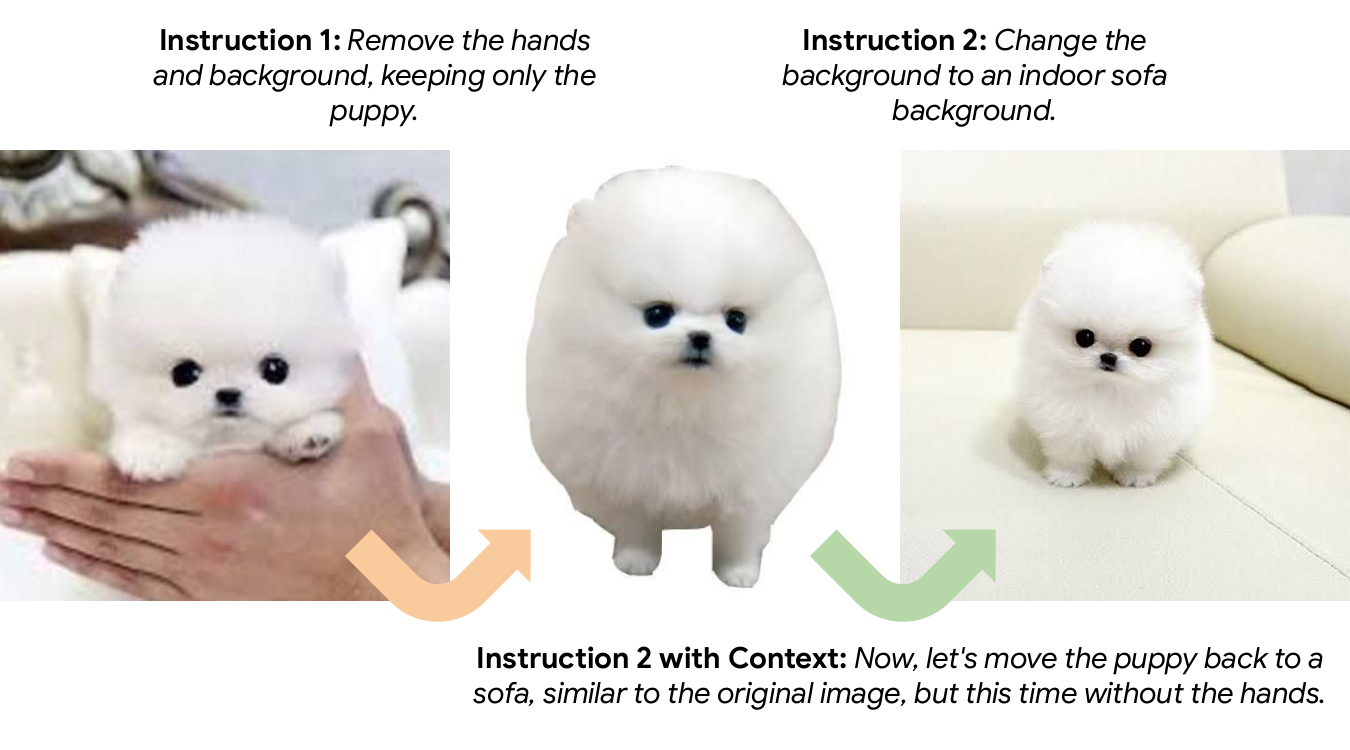}
        \caption{\textbf{Construction of multi-turn interaction chains.} An LMM rewrites subsequent instructions to be context-aware, transforming disjointed commands into a natural dialogue. The rewritten instruction references ``the original image'' and ``without the hands,'' and thus can test whether the model understands interaction history.}
        \label{fig:multi_turn}
    \end{minipage}
\end{figure*}

\noindent\textbf{4. Mining Hard Negatives.}
Effective contrastive learning training requires hard negatives, which are images that are semantically close to the target but do not perfectly satisfy the query. We incorporate a sophisticated, LMM-driven process for mining these negatives.
\begin{itemize}
    \item \textbf{For the Evaluation Set:} To maintain a static and unbiased benchmark, we first identify candidate negatives by finding the nearest neighbors of the source and target images within the entire web image source, using the feature space of the same pretrained image encoder from the first stage. We then use an LMM as a \emph{verifier}, which is prompted to perform a pairwise comparison between the ground-truth target and a candidate negative, judging which better fulfills the instruction while preserving the content of the source image (see an example in Fig.~\ref{fig:negative}). Only candidates judged to be inferior to the ground-truth target are included as hard negatives, ensuring that they are not false negatives. We collect multiple verified negatives for each query to enable more robust evaluation.
    \item \textbf{For the Training Set:} We employ a model-guided approach for augmenting the training data. We first train an initial \ourmodel model without hard negatives. This model is then used to retrieve the closest embeddings for each query, which serve as candidate hard negatives. These candidates are subsequently verified by the LMM using the same pairwise comparison logic. This creates a data engine that leverages the trained model's capabilities to discover progressively more challenging negatives for subsequent models to be trained. Due to computation limitations, we retain \emph{one hard negative} per training sample, and perform \emph{one iteration} of the model-guided hard negative mining.
\end{itemize}

\noindent\textbf{5. Constructing Multi-Turn Contextual Data.}
Finally, we compose the curated single-turn samples into multi-turn interaction chains. We identify sequences where the target image of one sample $(I_0 \rightarrow T_1 \rightarrow I_1)$ is the source image of another $(I_1 \rightarrow T_2 \rightarrow I_2)$, and concatenate them to form a multi-turn sequence $(I_0 \rightarrow T_1 \rightarrow I_1 \rightarrow T_2 \rightarrow I_2)$ (see an example in Fig.~\ref{fig:multi_turn}). We construct interactions of up to \emph{four turns}, because longer sequences are increasingly sparse. To make the dialogue more natural and context-aware, we prompt an LMM to rewrite the instructions for the second turn onwards ($T_2, \dots, T_m$). The rewritten instructions are designed to reference elements from the preceding context, transforming a series of disjointed commands into a coherent, interactive dialogue.

\noindent\textbf{6. Quality Control and Human Verification.}
Despite the automated nature of our data engine, the synthesized data maintains high quality due to our rigorous, multi-stage filtering process. To validate this, we conduct human verification on random samples from the evaluation data (representing 0.14\%, 0.96\%, 3.40\%, and 7.65\% of single-, two-, three-, and four-turn interactions, respectively) to ensure the retrieval trajectories are logical and correct. Among these samples, we observed an \emph{average validity of 90.5\%}, confirming the high quality of our evaluation set. Moreover, although our training dataset is fully synthetic, \ourmodel achieves state-of-the-art performance on established \emph{human-annotated benchmarks} (Tab.~\ref{tab:merged_results}) with $14\times$ fewer training samples than prior work~\cite{zhang2024magiclens}. This strong generalization demonstrates that our data engine successfully captures natural search behaviors rather than \emph{overfitting to synthetic artifacts or exhibiting self-preference bias}, as models trained on biased data would inherently struggle to transfer to external, human-curated domains.

%% file: sec/4_experiments.tex
\section{Experiments}
\label{sec:experiments}

In this section, we conduct a comprehensive evaluation of our proposed model. We first detail our experimental setup, and then assess our model's performance on established single-turn CIR benchmarks. Finally, we evaluate its core capability on our newly proposed \ourtask task. Implementation details, analysis of data efficiency, and full results on single-turn CIR are included in the supplementary material.

\subsection{Experimental Setup}
\label{sec:setup}

\noindent\textbf{Benchmarks.} We evaluate our model on three standard single-turn benchmarks: \textbf{Fashion IQ (FIQ)}~\cite{wu2021fashion}, \textbf{CIRR}~\cite{liu2021image}, and \textbf{CIRCO}~\cite{baldrati2023zero}. FIQ is a benchmark specialized for fashion retrieval; CIRR is a widely-used benchmark known for its realistic, open-ended human-generated instructions; CIRCO provides a large-scale, open-domain challenge, testing retrieval across a diverse set of concepts. To evaluate our primary contribution in the new multi-turn setting, we use our newly proposed \textbf{\ourtask benchmark}, which features dialogue-based image retrieval sequences of varying lengths (Sec.~\ref{sec:data}).

\noindent\textbf{Baselines.} We compare our approach against several state-of-the-art single-turn composed image retrieval methods. These include \textbf{MagicLens}~\cite{zhang2024magiclens}, which leverages synthetic data from generative models, \textbf{CIReVL}~\cite{karthik2024vision}, which uses context reasoning from foundation models, and other LMM-based embedding models like \textbf{E5-V}~\cite{jiang2024e5}. It is important to note that these baselines are \emph{not designed for multi-turn interactions} and architecturally cannot take multiple images as input. For the \ourtask benchmark, we need to apply some adaptation strategies (see details in Sec.~\ref{sec:multi_turn}) so that they can be comprehensively evaluated.

\subsection{CIR: Single-Turn Retrieval}
\label{sec:single_turn}

We first demonstrate our model's strong performance on the standard, single-turn Composed Image Retrieval (CIR) task. This evaluation indicates that our LMM-based architecture, even when fine-tuned for an interactive setting, excels at the fundamental single-turn retrieval problem. We evaluate on the FIQ, CIRR, and CIRCO benchmarks, using their standard evaluation protocols: FIQ and CIRR results are reported using Recall@$k$ metrics, while CIRCO results are reported using mean Average Precision (mAP).

As shown in Tab.~\ref{tab:merged_results}, our model establishes a new state of the art. On FIQ, we reach an overall Recall@10 of 40.1, and on CIRR, we achieve a Recall@1 of 38.7, significantly outperforming prior methods. Our performance on CIRCO is even more pronounced, with an mAP@5 of 39.4, showcasing our model's ability to handle fine-grained, open-domain instructions effectively. This strong single-turn performance serves as a crucial foundation, proving that our model's representations are robust before we even evaluate its more complex, multi-turn capabilities. Complete CIR results are included in the supplementary material.

\begin{table*}[t]
\centering
\caption{\textbf{Single-turn retrieval performance on standard benchmarks}. This evaluation validates that \ourmodel is a general and powerful retriever, even in a single-turn context. Our model achieves new state-of-the-art performance on CIRR and CIRCO and highly competitive results on FIQ (LinCIR's advantage on FIQ and weakness on CIRCO likely originate from a close linguistic alignment between its training corpus and FIQ; see the supplementary material). To provide a holistic performance comparison across metrics with varying scales, ``Rank $\downarrow$'' reports the average rank of each model across all six metrics.}
\label{tab:merged_results}
\resizebox{\textwidth}{!}{%
\begin{tabular}{@{}ll|cc|cc|cc|c@{}}
\toprule
\multirow{2}{*}{Model} & \multirow{2}{*}{Backbone} & \multicolumn{2}{c|}{FIQ} & \multicolumn{2}{c|}{CIRR} & \multicolumn{2}{c|}{CIRCO} & \multirow{2}{*}{Rank $\downarrow$} \\
& & R@10 & R@50 & R@1 & R@5 & mAP@5 & mAP@10 & \\
\midrule
CompoDiff~\cite{gu2024compodiff} & CLIP-G~\cite{radford2021clip} & 39.0 & 51.7 & 26.7 & 55.1 & 15.3 & 17.7 & 9.50 \\
E5-V~\cite{jiang2024e5} & LLaVA-NeXT-8B~\cite{li2024llavanext-strong} & 31.8 & 53.8 & 33.9 & 64.1 & 19.1 & 20.6 & 8.75 \\
MagicLens~\cite{zhang2024magiclens} & CLIP-L~\cite{radford2021clip} & 30.7 & 52.5 & 30.1 & 61.7 & 29.6 & 30.8 & 8.67 \\
SEARLE~\cite{baldrati2023zero} & CLIP-G~\cite{radford2021clip} & 34.8 & 55.7 & 34.8 & 64.1 & 13.2 & 13.9 & 7.92 \\
CIReVL~\cite{karthik2024vision} & CLIP-G~\cite{radford2021clip} & 32.2 & 52.4 & 34.7 & 64.3 & 26.8 & 27.6 & 7.83 \\
LDRE~\cite{yang2024ldre} & CLIP-G~\cite{radford2021clip} & 32.5 & 55.5 & 36.1 & 66.4 & 31.1 & 32.2 & 5.50 \\
MagicLens~\cite{zhang2024magiclens} & CoCa-L~\cite{yu2022coca} & 38.0 & 58.2 & 33.3 & 67.0 & 34.1 & 35.4 & 5.00 \\
LinCIR~\cite{gu2024language} & CLIP-G~\cite{radford2021clip} & \textbf{45.1} & \textbf{65.7} & 35.3 & 64.7 & 19.7 & 21.0 & 4.83 \\
BGE-VL~\cite{zhou2025megapairs} & CLIP-L~\cite{radford2021clip} & 34.6 & 55.2 & 38.0 & 70.3 & 39.2 & 40.2 & 3.75 \\
\textbf{\ourmodel-4B (Ours)} & Gemma3-4B~\cite{team2025gemma} & \underline{40.1} & 60.8 & \underline{37.7} & \underline{70.6} & \underline{37.8} & \underline{38.4} & \underline{2.75} \\
\textbf{\ourmodel-12B (Ours)} & Gemma3-12B~\cite{team2025gemma} & \underline{40.1} & \underline{61.0} & \textbf{38.7} & \textbf{70.8} & \textbf{39.4} & \textbf{40.2} & \textbf{1.50} \\
\bottomrule
\end{tabular}%
}
\end{table*}

\subsection{\ourtask: Multi-Turn Retrieval}
\label{sec:multi_turn}

We evaluate contextual retrieval on our proposed benchmark, which consists of dialogue chains varying from 1 to 4 turns. As defined in Sec.~\ref{sec:task}, we employ the $m$-turn Recall@$k$ metric. This metric is cumulative and unforgiving: it requires the model to successfully retrieve the correct target within the top-$k$ results at \textit{every single turn} of the sequence. A failure at any intermediate turn constitutes a failure for the entire session.

\begin{table*}[h!]
\centering
\caption{\textbf{Multi-turn retrieval performance on our \ourtask benchmark.} We report the $m$-turn Recall@$k$ for interactions up to 4 turns, where success requires correct retrieval at \emph{every} turn. Standard CIR models fail even when adapted with various strategies (including Gemini-summarized context). In contrast, our full-context model, \ourmodel, dramatically outperforms all baselines, proving that \ourmodel's end-to-end modeling is necessary for understanding multimodal context. \emph{Note:} Columns evaluate distinct, non-overlapping data splits, so metrics are not monotonic across turns.}
\label{tab:multi_turn}
\resizebox{\columnwidth}{!}{%
\begin{tabular}{@{}ll|cc|cc|cc|cc@{}}
\toprule
\multirow{2}{*}{Model} & \multirow{2}{*}{Backbone} & \multicolumn{2}{c}{1-turn} & \multicolumn{2}{c}{2-turn} & \multicolumn{2}{c}{3-turn} & \multicolumn{2}{c}{4-turn} \\
& & R@1 & R@5 & R@1 & R@5 & R@1 & R@5 & R@1 & R@5 \\
\midrule
\rowcolor{gray!25}\multicolumn{10}{@{}l}{\textit{First Image + Concatenated Instructions:} $(I_0, T_1+\dots+T_t)$} \\
\multicolumn{2}{@{}l|}{TIPS-SO400M~\cite{maninis2025tips}} & 25.2 & 51.0 & 0.4 & 21.3 & 0.0 & 11.0 & 0.0 & 5.1 \\
\multicolumn{2}{@{}l|}{SigLIP-SO400M~\cite{zhai2023sigmoid}} & 31.4 & 59.3 & 2.1 & 27.3 & 0.0 & 12.9 & 0.0 & 4.9 \\
\multicolumn{2}{@{}l|}{SigLIP2-SO400M~\cite{tschannen2025siglip}} & 31.0 & 59.2 & 1.6 & 29.4 & 0.0 & 12.1 & 0.0 & 3.6 \\
MagicLens~\cite{zhang2024magiclens} & CLIP-L~\cite{radford2021clip} & 36.7 & 57.8 & 8.1 & 26.7 & 1.0 & 11.4 & 0.0 & 2.3 \\
E5-V & LLaVA-NeXT-8B~\cite{li2024llavanext-strong} & 47.1 & 68.6 & 16.2 & 44.5 & 6.3 & 32.6 & 4.6 & 30.6 \\
BGE-VL-MLLM-S1 & LLaVA-NeXT-7B~\cite{liu2024llavanext} & 60.0 & 79.5 & 28.7 & 58.3 & 12.5 & 42.6 & 4.9 & 34.7 \\
\midrule
\rowcolor{gray!25}\multicolumn{10}{@{}l}{\textit{Last Image + Last Instruction:} $(I_{t-1}, T_t)$} \\
\multicolumn{2}{@{}l|}{TIPS-SO400M~\cite{maninis2025tips}} & 25.2 & 51.0 & 0.0 & 24.1 & 0.0 & 14.2 & 0.0 & 6.7 \\
\multicolumn{2}{@{}l|}{SigLIP-SO400M~\cite{zhai2023sigmoid}} & 31.4 & 59.3 & 0.0 & 31.5 & 0.0 & 18.6 & 0.0 & 11.0 \\
\multicolumn{2}{@{}l|}{SigLIP2-SO400M~\cite{tschannen2025siglip}} & 31.0 & 59.2 & 0.0 & 33.2 & 0.0 & 23.8 & 0.0 & 18.6 \\
MagicLens~\cite{zhang2024magiclens} & CLIP-L~\cite{radford2021clip} & 36.7 & 57.8 & 14.5 & 35.7 & 6.8 & 23.9 & 6.4 & 20.7 \\
E5-V & LLaVA-NeXT-8B~\cite{li2024llavanext-strong} & 47.1 & 68.6 & 19.2 & 46.7 & 15.6 & 38.8 & 13.4 & 37.2 \\
BGE-VL-MLLM-S1 & LLaVA-NeXT-7B~\cite{liu2024llavanext} & 60.0 & 79.5 & 36.2 & 64.1 & 28.6 & 51.8 & 27.0 & 54.0 \\
\midrule
\rowcolor{gray!25}\multicolumn{10}{@{}l}{\textit{Summarized Context:} $T'_t=\text{Gemini}(I_0, T_1,I_1,\dots,T_{t-1},I_{t-1},T_t)$} \\
\multicolumn{2}{@{}l|}{TIPS-SO400M~\cite{maninis2025tips}} & 25.7 & 51.9 & 10.6 & 22.7 & 5.2 & 11.8 & 1.2 & 5.9 \\
\multicolumn{2}{@{}l|}{SigLIP-SO400M~\cite{zhai2023sigmoid}} & 31.0 & 46.9 & 11.8 & 28.6 & 7.8 & 23.1 & 8.8 & 22.5 \\
\multicolumn{2}{@{}l|}{SigLIP2-SO400M~\cite{tschannen2025siglip}} & 51.6 & 71.4 & 23.3 & 48.7 & 18.0 & 40.9 & 16.7 & 33.7 \\
MagicLens~\cite{zhang2024magiclens} & CLIP-L~\cite{radford2021clip} & 43.5 & 66.5 & 19.8 & 47.0 & 12.7 & 37.8 & 10.7 & 33.0 \\
E5-V & LLaVA-NeXT-8B~\cite{li2024llavanext-strong} & 50.1 & 73.1 & 23.8 & 53.1 & 23.5 & 45.9 & 21.3 & 45.4 \\
BGE-VL-MLLM-S1 & LLaVA-NeXT-7B~\cite{liu2024llavanext} & 60.8 & 79.4 & 35.5 & 62.7 & 29.5 & 54.8 & 28.2 & 53.9 \\
\midrule
\rowcolor{cvprblue!25}\multicolumn{10}{@{}l}{\textit{Full Context:} $(I_0, T_1,I_1,\dots,T_{t-1},I_{t-1},T_t)$} \\
\textbf{\ourmodel-4B (Ours)} & Gemma3-4B~\cite{team2025gemma} & \underline{74.5} & \underline{89.8} & \underline{48.3} & \underline{79.5} & \underline{34.5} & \underline{69.3} & \underline{31.1} & \underline{65.8} \\
\textbf{\ourmodel-12B (Ours)} & Gemma3-12B~\cite{team2025gemma} & \bfseries 76.3 & \bfseries 90.4 & \bfseries 54.8 & \bfseries 82.1 & \bfseries 44.6 & \bfseries 76.6 & \bfseries 44.1 & \bfseries 78.3 \\
\bottomrule
\end{tabular}%
}
\end{table*}

\noindent\textbf{Baseline Adaptations.} Existing CIR models (\eg, MagicLens~\cite{zhang2024magiclens}) are architecturally limited to single-turn inputs (one reference image and one instruction). Because they inherently cannot process the multi-image, multi-turn history $H_m$, they cannot be directly evaluated on our benchmark. For a rigorous comparison, we build proxy inputs to adapt these single-turn models to the interactive setting using three distinct strategies (see visualization in the supplementary material):
\begin{enumerate}
    \item \textit{Concatenated Instructions:} We provide the initial source image ($I_0$) combined with the concatenation of all text instructions up to the current turn.
    \item \textit{Latest Inputs:} We pair the ground-truth image from the previous turn ($I_{t-1}$) with the current instruction ($T_t$) as the input.
    \item \textit{Summarized Context:} To give the baselines the strongest possible advantage, we use a state-of-the-art LMM (Gemini 2.5 Pro) as an oracle. We prompt the LMM to read the entire history up to the given step (including all intermediate images and text) and synthesize it into a single, optimized text instruction ($T'_t$) for the retrieval model. This isolates the model's retrieval capability by minimizing the requirement of complex context comprehension.
\end{enumerate}

\noindent\textbf{Results.} Tab.~\ref{tab:multi_turn} demonstrates the profound limitations of adapting single-turn models to multi-turn tasks and the clear superiority of our \emph{native end-to-end} framework. Standard baselines relying on the first two heuristic adaptations collapse rapidly as the interaction depth increases, falling to near-zero recall by the fourth turn. As expected, the \textit{Summarized Context} strategy acts as a strong oracle, significantly elevating the baseline performance. However, even when deliberately advantaged by Gemini 2.5 Pro's summarization capabilities, these baselines still underperform our method. For instance, our 12B model achieves a 4-turn R@1 of 44.1\%, whereas the best summarized baseline reaches only 28.2\%. This performance gap provides crucial evidence that multi-turn visual dialogue cannot be losslessly compressed into a single text prompt. \ourmodel's ability to natively ingest multiple images and internally maintain dialogue state allows it to handle multi-turn interactions, proving that end-to-end modeling is essential for contextual retrieval.

\subsection{Ablation Study}
\label{sec:ablation}

We conduct a comprehensive ablation study to validate the effectiveness of our data engine and model design choices and summarize results in Tab.~\ref{tab:ab-data} and \ref{tab:ab-model}. We analyze the data efficiency and scaling in the supplementary material.

\begin{table*}[t]
\centering
\begin{minipage}[b]{0.49\textwidth}
\centering
\caption{\textbf{Ablation study on the data source.} We compare the effectiveness of feature-based (Feat.) and metadata-based (Meta.) discovery strategies, and the impact of mining hard negatives (HN). Mixing sources and adding hard negatives yields the best performance. ``Rank $\downarrow$'' reports the average rank across all four metrics (lower is better).}
\label{tab:ab-data}
\resizebox{\textwidth}{!}{%
\begin{tabular}{@{}cc|ccccc@{}}
\toprule
\multirow{2}{*}{DS} & \multirow{2}{*}{HN} & FIQ & CIRR & CIRCO & \ourtask & \multirow{2}{*}{Rank $\downarrow$} \\
& & R@10 & R@1 & mAP@5 & 1-turn R@1 & \\
\midrule
Feat. & & 31.6 & 36.9 & 33.2 & 65.7 & 4.50 \\
Feat. & \ding{51} & 38.0 & 36.4 & \bf 38.0 & 74.5 & 2.75 \\
Meta. & & 34.2 & 36.6 & 32.7 & 60.7 & 4.75 \\
Meta. & \ding{51} & 38.3 & 34.3 & 33.8 & 56.4 & 4.50 \\
Mix & & 33.7 & \bf 38.0 & 34.8 & 67.7 & 3.00 \\
\rowcolor{cvprblue!25} Mix & \ding{51} & \bf 40.1 & 37.7 & 37.8 & \bf 75.4 & \bf 1.50 \\
\bottomrule
\end{tabular}%
}
\end{minipage}\hfill
\begin{minipage}[b]{0.49\textwidth}
\centering
\caption{\textbf{Ablation study on model design choices.} We analyze the impact of Bidirectional Attention (BA), Single-side Contrastive loss (SC), and the Special Embedding Token (ET). The full model combining all three designs achieves the highest accuracy. ``Rank $\downarrow$'' reports the average rank across all four metrics (lower is better).}
\label{tab:ab-model}
\resizebox{\textwidth}{!}{%
\begin{tabular}{@{}ccc|ccccc@{}}
\toprule
\multirow{2}{*}{BA} & \multirow{2}{*}{SC} & \multirow{2}{*}{ET} & FIQ & CIRR & CIRCO & \ourtask & \multirow{2}{*}{Rank $\downarrow$} \\
& & & R@10 & R@1 & mAP@5 & 1-turn R@1 & \\
\midrule
& \ding{51} & \ding{51} & 39.7 & 33.9 & 36.6 & 74.2 & 3.25 \\
\ding{51} & & \ding{51} & 39.6 & 35.9 & 37.4 & \bf 75.6 & 2.25 \\
\ding{51} & \ding{51} & & 39.2 & 37.2 & 35.9 & 75.5 & 3.00 \\
\rowcolor{cvprblue!25} \ding{51} & \ding{51} & \ding{51} & \bf 40.1 & \bf 37.7 & \bf 37.8 & 75.4 & \bf 1.50 \\
\bottomrule
\end{tabular}%
}
\end{minipage}
\end{table*}

\noindent\textbf{Data Source and Hard Negatives.} In Tab.~\ref{tab:ab-data}, we evaluate the impact of different data sources and the inclusion of hard negatives. We employ two strategies for discovering training pairs: feature-based similarity (Feat.) and metadata-based co-occurrence (Meta.). While both strategies independently yield strong results, mixing them in a balanced ratio provides a noticeable performance boost. This suggests that the two sources offer complementary signals, leading to greater diversity in the training data. Furthermore, the integration of LMM-verified hard negatives (HN) yields a substantial improvement on most benchmarks. This confirms that training with difficult negatives is crucial for forcing the model to learn fine-grained instruction following.

\noindent\textbf{Model Design Choices.} We further examine three key architectural designs in Tab.~\ref{tab:ab-model}: Bidirectional Attention (BA) within turns, the use of a Single-side Contrastive loss (SC), and the special \texttt{\textlangle EMB\textrangle} Token (ET). First, employing bidirectional attention proves superior to pure causal attention, as it allows for better information aggregation between image and text tokens within a specific turn. Second, we compare our objective function against a symmetric contrastive loss commonly used in CLIP training~\cite{radford2021clip}. We observe that the simpler, non-symmetric formulation (denoted as SC) works better for our retrieval task as it prioritizes distinction of candidates. Finally, using a dedicated \texttt{\textlangle EMB\textrangle} token (ET) to summarize the interaction history outperforms using the last token's hidden state, demonstrating the effectiveness of explicitly learning a compact context representation.

\begin{table}[ht]
\centering
\caption{\textbf{Ablation study on data scale.} We compare \ourmodel-4B trained on varying dataset sizes against the state-of-the-art MagicLens~\cite{zhang2024magiclens}. Remarkably, with just 320K samples (less than 1\% of the baseline's data size), our model already surpasses the baseline across all three single-turn CIR benchmarks. This underscores the high quality of our autonomously generated data and the data efficiency of our architecture. Scaling data size and model size together brings further performance gain (Table~\ref{tab:merged_results}).}
\label{tab:ab-scaling}
\begin{tabular}{@{}lc|ccc@{}}
\toprule
\multirow{2}{*}{Method} & \multirow{2}{*}{Data} & FIQ & CIRR & CIRCO \\
& & R@10 & R@1 & mAP@5 \\
\midrule
\rowcolor{gray!25} MagicLens CoCa-L~\cite{zhang2024magiclens} & 36.7M & 38.0 & 33.3 & 34.1 \\
\textbf{\ourmodel-4B (Ours)} & 320K & 38.5 & 34.8 & 36.7 \\
\textbf{\ourmodel-4B (Ours)} & 640K & 38.8 & 35.5 & \bf 38.7 \\
\textbf{\ourmodel-4B (Ours)} & 1.28M & 40.0 & 36.3 & 37.9 \\
\rowcolor{cvprblue!25} \textbf{\ourmodel-4B (Ours)} & 2.56M & \bf 40.1 & \bf 37.7 & 37.8 \\
\bottomrule
\end{tabular}
\end{table}

\noindent\textbf{Data Efficiency and Scale.} In Tab.~\ref{tab:ab-scaling}, we analyze the impact of training data size on model performance. Our approach is substantially more data-efficient than prior state-of-the-art methods. The strong baseline MagicLens~\cite{zhang2024magiclens} relies on a large synthetic dataset of 36.7M pairs to achieve its performance. In contrast, our \ourmodel-4B model already surpasses MagicLens on all three single-turn CIR benchmarks (FIQ~\cite{wu2021fashion}, CIRR~\cite{liu2021image}, and CIRCO~\cite{baldrati2023zero}) when trained on only 320K samples, corresponding to more than a $100\times$ reduction in data volume. As we further scale the training set to 640K and 1.28M samples, performance improves consistently, demonstrating the high quality and data efficiency of our autonomously generated data. We also observe that the gains begin to saturate beyond 1.28M samples for the 4B model, suggesting that performance is becoming bottlenecked by model capacity rather than data availability. This interpretation is consistent with Table~\ref{tab:merged_results}, where scaling from \ourmodel-4B to \ourmodel-12B yields further improvements. Taken together, these results suggest that our LMM-driven data engine provides strong and scalable training signals, while additional gains may be unlocked by pairing this data with larger models.

%% file: sec/5_conclusion.tex
\section{Conclusion}
In this work, we introduce \textbf{\ourtask}, a new task for multi-turn, contextual composed image retrieval that moves beyond the static, single-turn limitations of existing methods. We propose the \textbf{Transformable Image Embedding (\ourmodel)} model, an LMM-based architecture that processes the full dialogue history to generate evolving embeddings that capture the user's cumulative, iterative intent. A key enabler is our fully autonomous \textbf{data engine}, which leverages LMM-driven self-reflection and hard-negative verification to create the first large-scale, high-quality dataset for this task. Our experiments demonstrate \ourmodel's effectiveness. It achieves state-of-the-art performance on traditional single-turn benchmarks like CIRR and CIRCO, while also dramatically outperforming all baselines on our new multi-turn \ourtask task. Where prior methods collapse, \ourmodel's ability to understand multi-turn context allows it to handle complex, iterative refinement, providing a robust foundation for the next generation of interactive visual search.

\subsubsection*{Acknowledgements.}
This work was supported in part by NSF under Grants 2106825 and 2519216, the DARPA Young Faculty Award, the ONR Grant N00014-26-1-2099, and the NIFA Award 2020-67021-32799. 

%% file: sec/X_suppl.tex
\clearpage
\setcounter{page}{1}
\appendix

This supplementary material provides additional details and analyses that support the main paper. We first describe the training setup and the construction of our autonomously generated dataset, including the multi-turn evaluation split. Next, we report complete results on standard single-turn CIR benchmarks and provide a qualitative discussion of why native full-context modeling is essential for contextual retrieval. Finally, we include two complementary analyses on \ourtask: an alternative final-turn-only evaluation metric and a profiling study of inference cost across multiple turns, both of which further support the practicality and effectiveness of our approach.

\section{Model Training Details}

Our model is built upon the Gemma 3~\cite{team2025gemma} architecture with 4B parameters or 12B parameters, which utilizes the SigLIP-SO400M visual encoder~\cite{zhai2023sigmoid} and processes images at $896\times 896$ resolution. We initialize the model with Gemma 3 pretrained weights (except the final linear embedding projector), and fine-tune the model on our autonomously generated dataset for $10,000$ steps. We set $\tau=0.1$ and use a global batch size of $256$, resulting in a total of 2.56M training samples seen. The learning rate follows a linear warmup (for $1,000$ steps) and cosine annealing schedule, with a peak learning rate of $1.0 \times 10^{-4}$. We employ the Muon optimizer~\cite{jordan2024muon, liu2025muon}. The entire training process is completed in 4 hours on a pod of 64 TPU v5p accelerators.

\section{Dataset Construction Details}
In this section, we include details in our data construction procedure.

\noindent\textbf{Discovering Image Pairs with Transformations.} For feature-based discovery, we sample about 3.5M images from the WebLI dataset~\cite{chen2023pali}, cluster them into 36K groups based on representations produced by a CLIP-pretrained ResNet~\cite{radford2021clip, he2016deep}, and collect image pairs (up to 1,000 per cluster) whose feature similarities lie in the range of $[0.7, 0.97]$. For metadata-based discovery, we sample about 35M images from WebLI, group them by the URL information, and collect pairs (up to 1,000 per URL) whose feature similarities lie in a wider range of $[0.5, 0.97]$.

\noindent\textbf{Synthesizing Instructions with Self-Reflection.} We prompt Gemini 2.5 Flash~\cite{team2025gemini} to synthesize the instructions for image-pair transformations and reflect on 1) the similarity of the given image pair, and 2) the generated instruction. Gemini gives two scores (integers in the range of $[1, 5]$) to assess their quality, respectively, which are used for data filtering. We keep pairs with both scores equaling 5 in the evaluation set, and keep pairs whose two scores are at least 3 for training.

\noindent\textbf{Mining Hard Negatives.} We introduce hard negative samples to the evaluation set to create a challenging while meaningful benchmark. For each sample in the evaluation set, we use the same ResNet mentioned above to find the top-10 images that are most similar to the source image and the top-10 images closest to the target image, which are considered as candidates for hard negatives. We then prompt Gemini 2.5 Flash to compare the original target image and each hard negative candidate to determine which one better fits the transformation specified by the source image and instruction. If the candidate fails to match the original target, we consider it as a true negative sample and include it in the final candidate set for evaluation.
For the training set, as mentioned in the main paper, we first train a \ourmodel without hard negatives, and use it to locate the top-10 images whose embeddings are most similar to the query embeddings.

\noindent\textbf{Constructing Multi-Turn Data.} We search images which serve both as the source image in a transformation and the target image in another transformation, and join these two transformations into a multi-turn one. To encourage diversity, each image can be used for producing up to 3 multi-turn interactions. We then prompt Gemini 2.5 Flash to rewrite instructions (starting from the second turn) based on previous context.

\noindent\textbf{Evaluation Data Split by Interaction Turns.} Our evaluation benchmark is partitioned into four disjoint subsets according to the number of turns in the context. Specifically, the \ourtask benchmark contains 21,084 single-turn queries, 3,129 two-turn queries, 883 three-turn queries, and 392 four-turn queries. Each subset is evaluated independently, so the results across the 1-turn to 4-turn results are \emph{not} directly nested or expected to be monotonic. This design allows us to measure performance under progressively longer contexts while ensuring that each setting reflects a dedicated evaluation split.

\section{Complete Results on Single-Turn Composed Image Retrieval}
Due to limited space, we only include the performance of \ourmodel based on several major metrics on single-turn CIR benchmarks in the main paper. In Tabs.~\ref{tab:fiq}, \ref{tab:cirr}, and \ref{tab:circo}, we list the full evaluation results on the FIQ~\cite{wu2021fashion}, CIRR~\cite{liu2021image}, and CIRCO~\cite{baldrati2023zero} benchmarks, respectively. Overvall, our model \ourmodel demonstrates state-of-the-art performance on single-turn CIR benchmarks.

\begin{table*}[ht]
\centering
\caption{Full results on the FIQ benchmark.
}
\label{tab:fiq}
\resizebox{\columnwidth}{!}{%
\begin{tabular}{@{}ll|cc|cc|cc|cc@{}}
\toprule
\multirow{2}{*}{Model} & \multirow{2}{*}{Backbone} & \multicolumn{2}{c}{Dress} & \multicolumn{2}{c}{Shirt} & \multicolumn{2}{c}{Toptee} & \multicolumn{2}{c}{\textbf{Overall}} \\
& & R@10 & R@50 & R@10 & R@50 & R@10 & R@50 & R@10 & R@50 \\
\midrule
MagicLens~\cite{zhang2024magiclens} & CLIP-L~\cite{radford2021clip} & 25.5 & 46.1 & 32.7 & 53.8 & 34.0 & 57.7 & 30.7 & 52.5 \\
MagicLens~\cite{zhang2024magiclens} & CoCa-L~\cite{yu2022coca} & 32.3 & 52.7 & 40.5 & 59.2 & 41.4 & 63.0 & 38.0 & 58.2 \\
SEARLE~\cite{baldrati2023zero} & CLIP-G~\cite{radford2021clip} & 28.2 & 50.3 & 36.5 & 55.4 & 39.8 & 61.5 & 34.8 & 55.7 \\
CompoDiff~\cite{gu2024compodiff} & CLIP-G~\cite{radford2021clip} & 37.8 & 49.1 & 41.3 & 55.2 & 44.3 & 56.4 & 39.0 & 51.7 \\
CIReVL~\cite{karthik2024vision} & CLIP-G~\cite{radford2021clip} & 27.1 & 49.5 & 33.7 & 51.4 & 35.8 & 56.1 & 32.2 & 52.4 \\
LinCIR~\cite{gu2024language} & CLIP-G~\cite{radford2021clip} & 38.1 & 60.9 & 46.8 & 65.1 & 50.5 & 71.1 & \textbf{45.1} & \textbf{65.7} \\
LDRE~\cite{yang2024ldre} & CLIP-G~\cite{radford2021clip} & 26.1 & 51.1 & 35.9 & 58.6 & 35.4 & 56.7 & 32.5 & 55.5 \\
E5-V~\cite{jiang2024e5} & LLaVA-NeXT-8B~\cite{li2024llavanext-strong} & 23.8 & 47.5 & 36.4 & 56.4 & 35.3 & 57.5 & 31.8 & 53.8 \\
\textbf{\ourmodel-4B (Ours)} & Gemma3-4B~\cite{team2025gemma} & 34.8 & 56.8 & 44.0 & 63.0 & 41.5 & 62.5 & \underline{40.1} & 60.8 \\
\textbf{\ourmodel-12B (Ours)} & Gemma3-12B~\cite{team2025gemma} & 33.4 & 56.0 & 45.0 & 63.6 & 41.9 & 63.4 & \underline{40.1} & \underline{61.0} \\
\bottomrule
\end{tabular}%
}
\end{table*}

\begin{table*}[th]
\centering
\caption{Full results on the CIRR benchmark.}
\label{tab:cirr}
\begin{tabular}{@{}ll|ccc|ccc@{}}
\toprule
Model & Backbone & R@1 & R@5 & R@10 & R$_s$@1 & R$_s$@2 & R$_s$@3 \\
\midrule
MagicLens~\cite{zhang2024magiclens} & CLIP-L~\cite{radford2021clip} & 30.1 & 61.7 & 74.4 & 68.1 & 84.8 & 93.2 \\
MagicLens~\cite{zhang2024magiclens} & CoCa-L~\cite{yu2022coca} & 33.3 & 67.0 & 77.9 & \textbf{70.9} & \textbf{87.3} & \textbf{94.5} \\
SEARLE~\cite{baldrati2023zero} & CLIP-G~\cite{radford2021clip} & 34.8 & 64.1 & 75.1 & 68.7 & 84.7 & 93.2 \\
CompoDiff~\cite{gu2024compodiff} & CLIP-G~\cite{radford2021clip} & 26.7 & 55.1 & 74.5 & 64.5 & 82.4 & 91.8 \\
CIReVL~\cite{karthik2024vision} & CLIP-G~\cite{radford2021clip} & 34.7 & 64.3 & 75.1 & 68.0 & 84.9 & 93.2 \\
LinCIR~\cite{gu2024language} & CLIP-G~\cite{radford2021clip} & 35.3 & 64.7 & 76.1 & 63.4 & 82.2 & 92.0 \\
LDRE~\cite{yang2024ldre} & CLIP-G~\cite{radford2021clip} & 36.1 & 66.4 & 77.2 & \underline{68.8} & \underline{85.7} & 93.8 \\
E5-V~\cite{jiang2024e5} & LLaVA-NeXT-8B~\cite{li2024llavanext-strong} & 33.9 & 64.1 & 75.9 & - & - & - \\
\textbf{\ourmodel-4B (Ours)} & Gemma3-4B~\cite{team2025gemma} & \underline{37.7} & \underline{70.6} & \underline{81.1} & 67.3 & 84.9 & \underline{93.9} \\
\textbf{\ourmodel-12B (Ours)} & Gemma3-12B~\cite{team2025gemma} & \textbf{38.7} & \textbf{70.8} & \textbf{81.7} & 68.0 & 85.1 & 92.9 \\
\bottomrule
\end{tabular}
\end{table*}

\begin{table*}[th]
\centering
\caption{Full results on the CIRCO benchmark.}
\label{tab:circo}
\begin{tabular}{@{}ll|cccc@{}}
\toprule
Model & Backbone & mAP@5 & mAP@10 & mAP@25 & mAP@50 \\
\midrule
MagicLens~\cite{zhang2024magiclens} & CLIP-L~\cite{radford2021clip} & 29.6 & 30.8 & 33.4 & 34.4 \\
MagicLens~\cite{zhang2024magiclens} & CoCa-L~\cite{yu2022coca} & 34.1 & 35.4 & 38.1 & 39.2 \\
SEARLE~\cite{baldrati2023zero} & CLIP-G~\cite{radford2021clip} & 13.2 & 13.9 & 15.3 & 16.0 \\
CompoDiff~\cite{gu2024compodiff} & CLIP-G~\cite{radford2021clip} & 15.3 & 17.7 & 19.4 & - \\
CIReVL~\cite{karthik2024vision} & CLIP-G~\cite{radford2021clip} & 26.8 & 27.6 & 30.0 & 31.0 \\
LinCIR~\cite{gu2024language} & CLIP-G~\cite{radford2021clip} & 19.7 & 21.0 & 23.1 & 24.2 \\
LDRE~\cite{yang2024ldre} & CLIP-G~\cite{radford2021clip} & 31.1 & 32.2 & 35.0 & 36.0 \\
E5-V~\cite{jiang2024e5} & LLaVA-NeXT-8B~\cite{li2024llavanext-strong} & 19.1 & 20.6 & 23.1 & 24.0 \\
\textbf{\ourmodel-4B (Ours)} & Gemma3-4B~\cite{team2025gemma} & \underline{37.8} & \underline{38.4} & \underline{40.9} & \underline{41.9} \\
\textbf{\ourmodel-12B (Ours)} & Gemma3-12B~\cite{team2025gemma} & \textbf{39.4} & \textbf{40.2} & \textbf{42.7} & \textbf{43.8} \\
\bottomrule
\end{tabular}
\end{table*}

We note that LinCIR's advantage on FIQ and weakness on CIRR and CIRCO likely originate from a close linguistic alignment between its training corpus and FIQ. To investigate this, we use a text embedding model Qwen3-Embedding~\cite{qwen3embedding} to encode the training samples of LinCIR and relative captions in the single-turn benchmarks. For each benchmark sample, we retrieve the nearest training sample and record its cosine similarity. Taking the maximum similarity across the entire training pool, 66.9\% of FIQ samples exceed a similarity of 0.85, compared to only 43.6\% for CIRCO and 23.3\% for CIRR. Some pairs are semantically close or even identical because FIQ contains many queries of simple words (\eg, ``floral design,'' ``more red''). This confirms that LinCIR's training distribution is substantially close to FIQ's retrieval queries, inflating its performance on FIQ while understating its generalization gap on CIRR and CIRCO.

\section{Context Encoding in Multi-Turn Retrieval}

\begin{figure}[!t]
    \centering
    \includegraphics[width=\columnwidth]{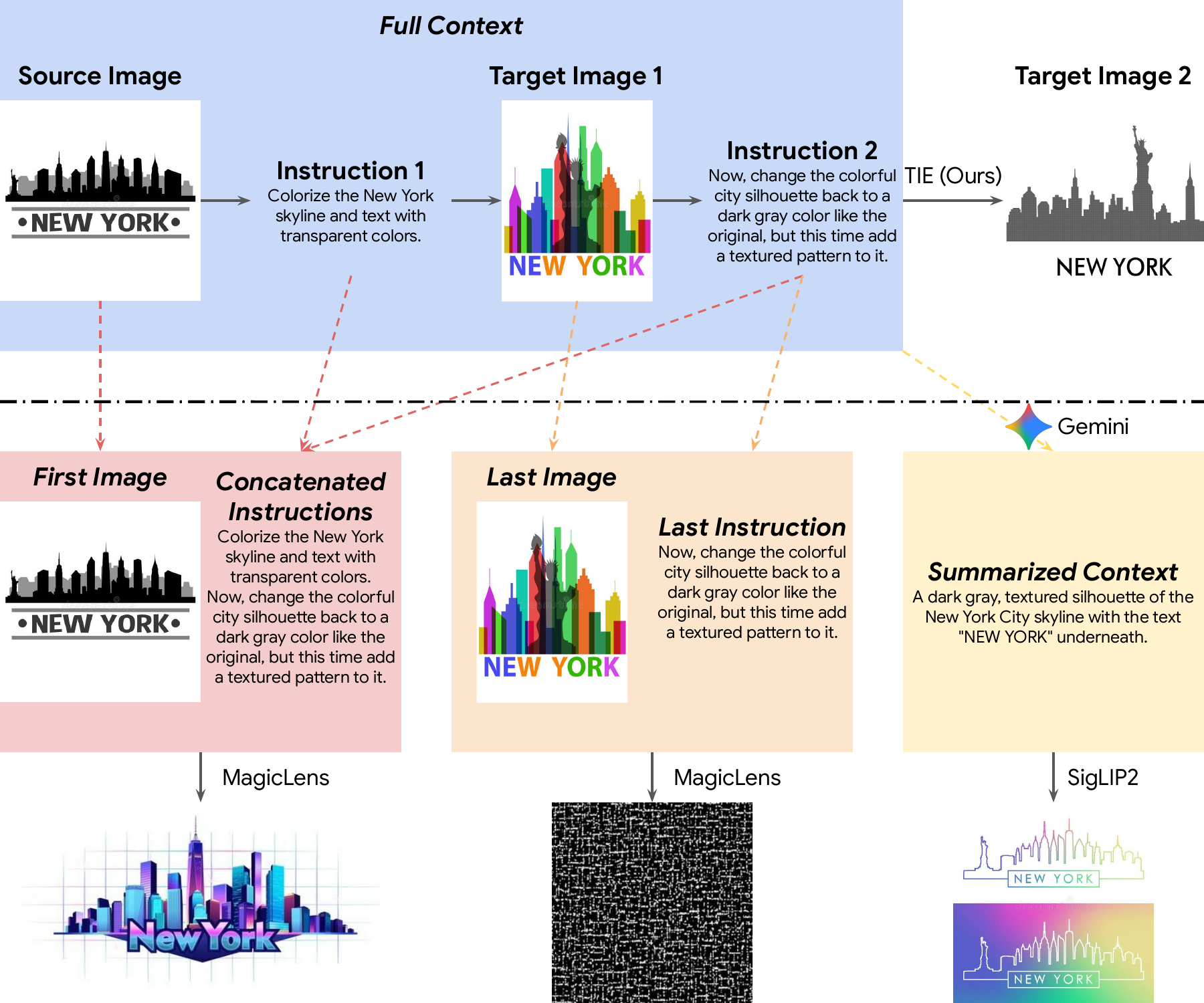}
    \caption{\textbf{Comparison of context encoding strategies for \ourtask.} The top illustrates our \emph{Full Context} approach (\ourmodel), which encodes the complete sequence of images and instructions, allowing it to correctly interpret context-dependent references (\emph{e.g.}, ``like the original''). The bottom depicts three adaptation strategies for single-turn baselines: (1) \emph{First Image + Concatenated Instructions}, which combines all text but ignores intermediate visual states; (2) \emph{Last Image + Last Instruction}, which uses the immediate predecessor but loses access to earlier history (failing the ``back to... original'' command); and (3) \emph{Summarized Context}, which uses an external LMM to rewrite the history into a single prompt. As shown, only \ourmodel successfully retrieves the correct target by preserving the full dialogue context.}
    \label{fig:context}
\end{figure}

The key limitation of prior single-turn CIR models is their fixed input architecture, which is strictly designed to process at most one reference image and one instruction at a time. They fundamentally lack the capacity to process a variable-length sequence of multiple images and texts. Consequently, directly feeding the full history is architecturally impossible, necessitating the adaptation strategies (see Sec.~\ref{sec:experiments}) to compress the multi-turn context into a compatible single-turn format.

Fig.~\ref{fig:context} visually demonstrates why adapting single-turn models is insufficient for \ourtask. We compare our native multi-turn approach \ourmodel against three adaptation strategies defined in our experimental setup.

The \emph{First Image + Concatenated Instructions} strategy (bottom left) overwhelms the model with potentially conflicting textual commands while discarding critical intermediate visual states. The \textit{Last Image + Last Instruction} strategy (bottom middle), which pairs the latest target with the current instruction, suffers from a loss of context; as illustrated, the instruction ``change... back to... like the original'' becomes impossible to resolve because the model has no access to the original source image. Finally, while the \emph{Summarized Context} strategy (bottom right) utilizes an external LMM to compress history into a single description, it acts as an imperfect proxy, often losing fine-grained visual details required for precise retrieval. In contrast, our model \ourmodel (top) encodes the entire multimodal history, successfully maintaining the dialogue context to resolve complex, recursive references.

\section{Final-Turn Recall as an Alternative Evaluation Metric}

We define $m$-turn Recall@$k$ as a metric that requires successful retrieval at \emph{every} turn in the context. As a complementary analysis, we also report an alternative metric that evaluates only the retrieval accuracy at the \emph{final} turn of each interaction. Concretely, for an $m$-turn interaction, final-turn Recall@$k$ checks whether the ground-truth target image at turn $m$ is contained in the model's top-$k$ predictions at that final step, regardless of whether earlier turns were successful. This metric relaxes the cumulative requirement and focuses solely on the endpoint of the interaction. As shown in Tab.~\ref{tab:multi_turn_final_only}, our \ourmodel model remains clearly stronger than the strongest baseline under this more permissive evaluation protocol, indicating that the advantage of end-to-end contextual modeling does not depend on the stricter trajectory-level metric.

\begin{table*}[th]
\centering
\caption{\textbf{Final-turn retrieval performance on \ourtask.} As an alternative to $m$-turn Recall@$k$, we report a relaxed version of Recall@$k$ computed only on the final turn of each interaction. Even under this more permissive metric, our full-context model still substantially outperforms the summarized-context baseline. \emph{Note:} Columns evaluate distinct, non-overlapping data splits grouped by interaction turns.}
\label{tab:multi_turn_final_only}
\resizebox{\columnwidth}{!}{%
\begin{tabular}{@{}ll|cc|cc|cc|cc@{}}
\toprule
\multirow{2}{*}{Model} & \multirow{2}{*}{Backbone} & \multicolumn{2}{c}{1-turn} & \multicolumn{2}{c}{2-turn} & \multicolumn{2}{c}{3-turn} & \multicolumn{2}{c}{4-turn} \\
& & R@1 & R@5 & R@1 & R@5 & R@1 & R@5 & R@1 & R@5 \\
\midrule
\rowcolor{gray!25}\multicolumn{10}{@{}l}{\textit{Summarized Context:} $T'_t=\text{Gemini}(I_0, T_1,I_1,\dots,I_{t-1},T_t)$} \\
BGE-VL-MLLM-S1 & LLaVA-NeXT-7B~\cite{liu2024llavanext} & 60.8 & 79.4 & 47.2 & 71.6 & 38.9 & 63.5 & 35.1 & 59.8 \\
\midrule
\rowcolor{cvprblue!25}\multicolumn{10}{@{}l}{\textit{Full Context:} $(I_0, T_1,I_1,\dots,I_{t-1},T_t)$} \\
\textbf{\ourmodel-4B (Ours)} & Gemma3-4B~\cite{team2025gemma} & \textbf{74.5} & \textbf{89.8} & \textbf{61.3} & \textbf{84.7} & \textbf{52.8} & \textbf{78.9} & \textbf{48.6} & \textbf{74.2} \\
\bottomrule
\end{tabular}%
}
\end{table*}

\section{Inference Cost Across Multiple Turns}

To understand the computational overhead of modeling long multimodal context, we perform a simulated profiling study on TPU devices and measure the peak HBM usage and total FLOPs under different numbers of turns and total text sequence lengths. The reported \emph{text sequence length} includes all text tokens and special tokens across each turn, but excludes the image tokens produced by the visual encoder (each image contributes a fixed number of 256 visual tokens). The reported HBM usage also includes model parameters (9.47 GiB for the 4B checkpoint; most parameters in bf16 and a small portion of the vision encoder in fp32).

\noindent\textbf{Scaling with the number of turns.}
Both memory usage and computational cost grow approximately linearly with the number of turns. The image-only configuration uses about 10.09 GiB of HBM, and each additional turn increases memory by roughly 1.1 GiB. Similarly, FLOPs increase steadily with the number of turns; for example, at a text sequence length of 128 tokens, computation rises from 8.0T FLOPs at one turn to 15.3T, 22.6T, and 29.9T FLOPs at two, three, and four turns respectively. This indicates that the dominant cost of modeling longer contexts comes from incorporating additional visual context.

\noindent\textbf{Effect of text sequence length.}
In contrast, increasing the number of text tokens within the same number of turns has a much smaller effect on memory usage. For example, with one turn, increasing the sequence length from 128 to 2,048 tokens changes peak HBM only slightly (10.09 GiB). However, FLOPs increase more noticeably because attention scales with sequence length; at one turn, the cost rises from 8.0T FLOPs (128 tokens) to 21.9T FLOPs (2,048 tokens). The same trend holds for multi-turn settings.

\noindent\textbf{Summary.}
Overall, the profiling results show that the system cost scales predictably with the number of turns. Adding more turns primarily increases both memory and compute due to additional visual context, while increasing the text sequence length mainly affects computation but has minimal impact on memory. This scaling behavior suggests that modeling multi-turn multimodal context with full context remains efficient on modern accelerators.

\begin{table}[t]
\centering
\caption{\textbf{Representative TPU profiling results.} High Bandwidth Memory (HBM) utilization includes model parameters (9.47 GiB). Text sequence length counts all text and special tokens but excludes image tokens.}
\label{tab:profiling_summary}
\begin{tabular}{@{}c|c|c|c@{}}
\toprule
\# Turns & Text Seq Len & Peak HBM (GiB) & FLOPs (T) \\
\midrule
0 & 16 & 10.09 & 7.30 \\
\midrule
1 & 128 & 10.09 & 8.04 \\
1 & 2,048 & 10.09 & 21.87 \\
\midrule
2 & 128 & 11.17 & 15.29 \\
2 & 2,048 & 11.17 & 29.40 \\
\midrule
4 & 128 & 13.30 & 29.91 \\
4 & 2,048 & 13.30 & 44.58 \\
\midrule
6 & 128 & 15.46 & 44.91 \\
6 & 2,048 & 15.48 & 60.14 \\
\bottomrule
\end{tabular}
\end{table}

\section{Limitations and Future Work}

Despite the strong performance of our approach, we identify some limitations of our current benchmark and model. First, the most prominent failure mode is \textbf{fine-grained spatial understanding}: the model can struggle with transformations involving precise object layouts, orientations, relative positions, or localized edits, especially when these spatial relations must be tracked across multiple turns. We expect that incorporating more samples with corrected spatial understanding can mitigate this issue. Second, although our data engine produces high-quality large-scale training data, it remains largely synthetic and may not fully capture the full diversity of real interactive retrieval behaviors. Third, our evaluation benchmark currently contains a limited number of long interactions and maximum total turns, which limits a more comprehensive analysis of very deep interaction histories. We view richer spatially grounded data, more human-authored interactive retrieval trajectories, and more challenging long-horizon benchmarks as important directions for future work.